\documentclass[conference]{IEEEtran}
\usepackage{enumerate}
\usepackage{graphicx}
\usepackage{epsf}
\usepackage{subfigure}
\usepackage{amsmath}
\usepackage{amssymb}
\usepackage{array}
\usepackage{setspace}
\usepackage[amsmath,thmmarks]{ntheorem}
\usepackage[ruled,lined]{algorithm2e}
\usepackage{algorithmic}
\usepackage{color}
\usepackage{amsfonts}
\usepackage{dsfont}
\usepackage{xfrac}
\usepackage{breqn}
\usepackage{bbm}
\usepackage{cite}

\usepackage{diagbox}
\usepackage{multirow} 

\graphicspath{{Fig/},{fig/}}

\theoremseparator{.}

\begin{document}
\title{\huge Hierarchical Multi-Agent Multi-Armed Bandit for Resource Allocation in Multi-LEO Satellite Constellation Networks}

\author{\IEEEauthorblockN{Li-Hsiang Shen, Yun Ho, Kai-Ten Feng, Lie-Liang Yang$^*$, Sau-Hsuan Wu, and Jen-Ming Wu$^{\dagger}$}
\IEEEauthorblockA{\footnotesize
Department of Electronics and Electrical Engineering, National Yang Ming Chiao Tung University, Hsinchu, Taiwan\\
*Next Generation Wireless, University of Southampton, Southampton, UK\\
$^{\dagger}$Next-generation Communications Research Center, Hon Hai Research Institute, Taipei, Taiwan\\
Email: gp3xu4vu6.cm04g@nctu.edu.tw, yunho.ee10@nycu.edu.tw, ktfeng@nycu.edu.tw, \\ lly@ecs.soton.ac.uk, sauhsuan@gmail.com, and jen-ming.wu@foxconn.com
%\IEEEauthorrefmark{1}phtseng@mail.moj.gov.tw
}}

%% define terminology %%
\newcommand{\SYS}{IPF\xspace}

\maketitle
%\footnotetext[1] {This work was in part funded by MOST 107-2622-8-009-020, 107-2218-E-009-047, 107-2221-E-009-058-MY3,  107-2221-E-027-040-MY2, and the Wistron NeWeb Corporation (WNC), Taiwan.}

\begin{abstract}
%In recent years, wireless signal-based vital signs monitoring techniques have attracted significant attention. 
Low Earth orbit (LEO) satellite constellation is capable of providing global coverage area with high-rate services in the next sixth-generation (6G) non-terrestrial network (NTN). Due to limited onboard resources of operating power, beams, and channels, resilient and efficient resource management has become compellingly imperative under complex interference cases. However, different from conventional terrestrial base stations, LEO is deployed at considerable height and under high mobility, inducing substantially long delay and interference during transmission. As a result, acquiring the accurate channel state information between LEOs and ground users is challenging. Therefore, we construct a framework with a two-way transmission under unknown channel information and no data collected at long-delay ground gateway. In this paper, we propose hierarchical multi-agent multi-armed bandit resource allocation for LEO constellation (mmRAL) by appropriately assigning available radio resources. LEOs are considered as collaborative multiple macro-agents attempting unknown trials of various actions of micro-agents of respective resources, asymptotically achieving suitable allocation with only throughput information. In simulations, we evaluate mmRAL in various cases of LEO deployment, serving numbers of users and LEOs, hardware cost and outage probability. Benefited by efficient and resilient allocation, the proposed mmRAL system is capable of operating in homogeneous or heterogeneous orbital planes or constellations, achieving the highest throughput performance compared to the existing benchmarks in open literature.
\end{abstract}

\begin{IEEEkeywords}
Resource allocation, LEO constellation, unknown wireless channel, multi-agent, multi-armed bandit, machine learning.
\end{IEEEkeywords}

\section{Introduction}
Low Earth orbit (LEO) satellite constellation has been developed rapidly, bringing a large number of prospects in sixth-generation (6G) non-terrestrial network (NTN) communications \cite{leo1, leo2, shen1}. Lots of global organization and industry, such as SpaceX, Amazon, OneWeb, and TeleSAT have announced investment for extensive services and advanced researches to provide global-coverage applications \cite{hanzo1, shen1, leo2}. The NTN specification is specified by 3rd Generation Partnership Project (3GPP) since Release 15, aiming for global-coverage and throughput-aware services \cite{leo2}. 
%Later Releases from 16--18 for new radio (NR) additionally leverage space-air-terrestrial-maritime networks for supporting latency, reliability, massive connection-oriented services. 
With increasing demands and diverse services, several other projects like SAT5G and SATis5 are involving in NTN technologies, especially in LEOs \cite{leo1}.

LEO satellites possess several benefits compared to the conventional satellite communications of medium Earth orbit (MEO), high Earth orbit (HEO) and geostationary orbit (GEO) \cite{shen1}. Comparatively, lower LEO height allows lower latency, alleviated pathloss and lower deployment cost. To conquer the huge pathloss compared to terrestrial channels, LEOs have to conduct multiple narrow beamforming to concentrate power with high beam-gain to the desired users in several directions \cite{multibeam}. Moreover, massive LEOs are deployed and controlled by the ground gateway to support seamless coverage not covered by terrestrial infrastructure \cite{RABPC, LEOpromising2}. Nevertheless, under such dense LEO networks, limited resources should be appropriately assigned to prevent severe backdrops of inter-LEO, inter-beam and Doppler-induced interferences from high-mobility \cite{interf, interf2, LEOmovement}, which becomes compellingly imperative.

There exists abundant research results on different aspects of LEO networks. In \cite{TurnoffBeam}, they adaptively turn off the unnecessary beams to avoid inter-beam interference to conserve onboard power. The work of \cite{RAbeampower, interf2} jointly considers LEO beam-hopping and power allocation optimization. The authors in \cite{RABPC} have taken into account channel selection and beam hopping as well as power allocation to fulfill traffic requirement. In \cite{ComLett22}, they design bandwidth assignment while guaranteeing user rates under a multi-beam LEO system. The above-mentioned solutions of \cite{TurnoffBeam,RAbeampower, interf2, RABPC, ComLett22} are conducted in a centralized manner via ground gateway leading to high latency. The channel information is also required as the input data, which is impractical due to massive connections and long-distance. Additionally, they did not consider mobility and complicated interferences among beams and inter-satellites.

To resolve the above-mentioned problems, an action-reward based mechanism has been regarded as a promising solution. Without requiring channel information, reinforcing mechanisms are candidate solutions. The agents may interact with the environment to observe the reward for the following actions. In \cite{RAP1}, they perform power optimization with multi-agent beams by adopting reinforcement learning but with additional state data collection. The work in \cite{RAP2} has devised a game-based approach for allocating bandwidth to multiple beam-agents but with local solutions. In \cite{Coor1, Coor2}, collaborative LEO tasks are scheduled in different clusters; however, ground gateway should gather raw data from LEOs provoking considerable computing and delay. While, in \cite{Coor3}, MEO is considered as the edge collecting data from LEOs, which has insufficient processing capability and high-latency problem. In this context, a plethora of LEO works in \cite{RAP1, RAP2, Coor1, Coor2, Coor3} requires additional gateway for gathering impractically heavy data loads.

Therefore, leveraging multi-agent \cite{benchmark2} and multi-armed bandit (MAB) \cite{mab1} becomes the potential candidate to dynamically adjust policies with combinational environmental results. The notion of MAB includes exploitation and exploration, where the next action is determined randomly or based on the historically-accumulated rewards \cite{mab2}. Different from state-driven reinforcement learning \cite{benchmark3, benchmark4}, MAB only needs actions and rewards, which reduces additional overhead of learning. Moreover, multi-agent MAB is capable of learning in an online manner without an edge controller for highly-mobile LEOs. Due to above-mentioned backdrops, we conceive an enhanced multi-agent MAB based LEO architecture considering comprehensive resource allocation, different interference types as well as framework without channel estimation, which are not jointly considered in existing literature of \cite{TurnoffBeam,RAbeampower, interf2, RABPC, ComLett22, RAP1, RAP2, Coor1, Coor2, Coor3, benchmark2, benchmark3, benchmark4}. The main contributions of this paper are summarized as follows.
\begin{itemize}
    \item We conceive a multi-LEO constellation system with two-way transmission framework without explicitly utilizing channel information. Therefore, the ground gateway is not further required for computing optimization, leading to comparably lower overhead of data collection. We also consider the comprehensive interference factors of inter-LEO, inter-beam and Doppler-induced interferences.
    
    \item We propose a multi-agent multi-armed bandit resource allocation scheme for LEO constellation (mmRAL) by learning allocation of available power, beams and channel resources. Hierarchical agents are designed, including collaborative multiple macro-agents of LEOs, and competitive micro-agents of respective resources, i.e., power/beam/channel-agents. Greedy-based exploration is performed to prevent local solutions. The proposed mmRAL scheme asymptotically obtains suitable allocation with only throughput feedbacks.

    \item In simulations, we evaluate mmRAL in various cases of LEO deployment, numbers of serving users and operating LEOs, hardware cost as well as outage probability. Benefited by efficient and resilient allocation, the proposed mmRAL scheme is capable of operating in both homogeneous and heterogeneous orbital planes or constellations, achieving the highest throughput compared to the existing benchmarks in open literature.     
\end{itemize}

The rest of this paper is organized as follows. Section \ref{SM} describes the LEO system model and problem formulation. Section \ref{alg} elaborates the proposed mmRAL scheme, whilst  Section \ref{sim} shows substantial performance evaluation. Finally, the conclusions are drawn in Section \ref{con}.

\section{System Model and Problem Formulation} \label{SM}
\begin{figure}
  \centering
  \includegraphics[width=3.4in]{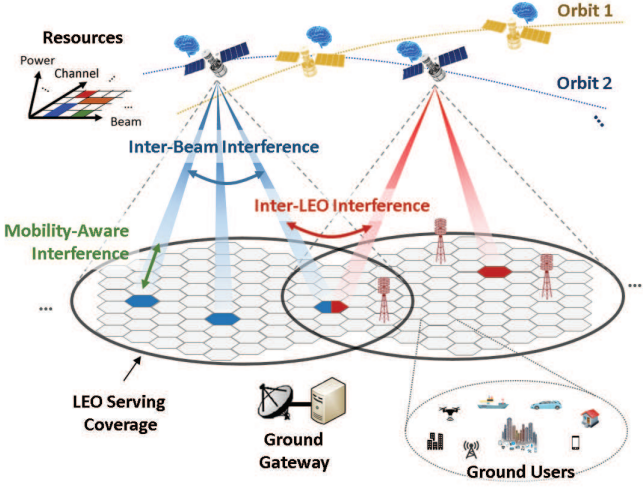}
  \caption{System architecture of heterogeneous multi-LEO satellite constellation networks. All operated localized LEOs are synchronized by a ground gateway serving multiple ground users. Note that the satellites with same color of orbits move within the homogeneous orbital plane. There induce several interferences, including inter-beam, inter-LEO and mobility-aware interference.}
  \label{fig:system model}
\end{figure}
%講目的
%實驗示意圖,說明

As shown in Fig. \ref{fig:system model}, we consider a heterogeneous LEO constellation, consisting of $N$ LEOs serving total $U$ randomly distributed ground users. We assume that LEOs are deployed with identical heights of $h_{n}$, $\forall 1\leq n \leq N$. The LEO constellation is operating at center frequency of $f_c$. Each LEO has the equal number of beam cells $M$ and total number of $S$ sub-channels under orthogonal frequency division multiplexing (OFDM) technique. Each sub-channel has $W_s$ bandwidth with total bandwidth of $W=\sum_{s=1}^S W_s$. For simplicity, we define $\kappa_{n,m,u}=1$ indicating that the $u$-th ground user is geometrically located in the $m$-th cell served by the $n$-th LEO under assumption of periodic location feedbacks. Note that a single ground gateway is deployed to be connected to LEOs, which is responsible for initial signalling. The LEOs are operated at Ka-band with each beam cell operating at frequency $f_{n,m}$. We model the asymptotic beamforming gain as
\begin{equation} \label{beamgain}
G\left(\theta,\psi^{(bs)}\right) =\left\{  
\begin{aligned}  
& G_t,\quad \text{if } \theta = \psi^{(bs)} ,\\  
& G_t \cdot4\left| \frac{J_1\left(ka \cdot\sin\left(\theta-\psi^{(bs)}\right)\right)}{ka\cdot \sin\left(\theta-\psi^{(bs)}\right)}\right|^2,       \\ & \qquad\qquad \text{if }  0 < \lvert\theta-\psi^{(bs)}\rvert \le \theta^{(max)}, \\ 
\end{aligned}  
\right.
\end{equation}
where $G_t$ is the transmit antenna gain, $\psi^{(bs)}$ is the given boresight angle that the generated beam is pointing at, depending on the geometric location of LEO and configured beam cells. $J_1(x)$ is the Bessel function of the first kind and first order with argument $x$, $a$ is radius of antenna aperture, $k = \pi /c$ is the number of waves, $c$ is the speed of light, and $\theta$ is the angle measured from the boresight. Note that the required number of antennas can be derived by \cite{doppler}. Based on $\eqref{beamgain}$, the signal-to-interference-plus-noise-ratio (SINR) of the $s$-th sub-channel for the $u$-th user served by the $m$-th beam and the $n$-th LEO can be acquired as
\begin{align} \label{SINR}
	& \Gamma_{n,m,u,s} = \nonumber \\ &
\frac{P_{n,m,u,s} G^T_{n,m,u}\left(\theta_{n,m,u}, \psi_{n,m,u}^{(bs)}\right) \cdot G^R_u H_{n,m,u,s} \phi_{n,m,u} \rho_{n,m,s}}{I_{B} + I_S + \eta \cdot I_{D} + \sigma^2},
\end{align}
where $P_{n,m,u,s}$, $G^T_{n,m,u}\left(\theta_{n,m,u}, \psi_{n,m,u}^{(bs)}\right)$, $G^{R}_{u}$, $H_{n,m,u,s}$, $\phi_{n,m,u}\in \left\{ 0,1 \right\}$, and $\rho_{n,m,s}\in \left\{ 0,1 \right\}$ respectively indicate the transmit power, transmit and receiving beam gains derived from $\eqref{beamgain}$, channel gain, and beam indicator as well as sub-channel allocation indicator of LEO $n$ in cell $m$ serving user $u$ using sub-channel $s$. The receiver gain $G^R_u$ is assumed to be constant due to its smaller scale of user's antenna orientation compared to LEOs. Note that channel $H_{n,m,u,s} = 10^{-PL/10}$ is related to sub-channel frequency $f_{n,m,s}$ and distance $d_{n,m,u}$, where $PL (\text{dB}) = PL_b + PL_g + PL_s$. $PL_b$ denotes basic pathloss including free space pathloss, shadow fading and clutter loss. $PL_g$ is atmospheric absorption loss, whilst $PL_s$ comes from tropospheric scintillation effects. The pertinent expressions and lookup table can be found in \cite{channel1, channel2}. The beam allocation $\phi_{n,m,u}=1$ when user $u$ is served by \textit{illuminated} beam $m$ under LEO $n$, whilst $\rho_{n,m,s}=1$ when LEO $n$'s beam $m$ is operated at channel $s$. Let us study the interference terms in $\eqref{SINR}$ as follow. The first term of denominator in $\eqref{SINR}$ stands for inter-beam interference (IBI) of intra-LEO, given by
\begingroup
\allowdisplaybreaks
\begin{align} \label{ibi}
	I_{B} &= \sum_{u'\neq u}^{U} \sum_{m'\neq m}^{M} P_{n,m',u',s} G_{n,m',u'}^T\left(\theta_{n,m',u'},\phi_{n,m,u}^{(bs)} \right) \nonumber \\ &\qquad\qquad\qquad\qquad \cdot G^R_uH_{n,m,u,s} \phi_{n,m',u'} \rho_{n,m',s}.
\end{align}
\endgroup
The inter-satellite interference (ISI) among LEOs is formulated as
\begingroup
\allowdisplaybreaks
\begin{align}
I_{S} & = \sum_{n'\neq n}^{N} \sum_{u'\neq u}^{U} \sum_{m'\neq m}^{M} P_{n',m',u',s} G_{n',m',u'}^T \left(\theta_{n',m',u'},\phi_{n,m,u}^{(bs)} \right) \nonumber \\ &\qquad\qquad\qquad\qquad\qquad \cdot G^R_uH_{n,m,u,s} \phi_{n',m',u'} \rho_{n',m',s}.
\end{align}
\endgroup
The Doppler-induced interference of LEOs can be asymptotically derived from \cite{doppler} as
\begingroup
\allowdisplaybreaks
\begin{align}
I_{D} \approx \frac{1}{2S} \left( \frac{f_{n,m,s} \cdot v_{n,u} S^2}{c \cdot W} \right)^2  \cdot \sum_{s=1}^S \sum_{s'\neq s}^S \frac{P_{n,m,u,s'}}{\left(s'-s \right)^2},
\end{align}
\endgroup
where $v_{n,u}$ is the relative velocity between LEO $n$ and user $u$. In $\eqref{SINR}$, we define $0\leq \eta < 1$ as the Doppler shift compensation constant. The final term in $\eqref{SINR}$ is sub-channel white noise power $\sigma^2 = N_{0}W_s$, where $N_{0}$ is noise power spectral density. Based on SINR in $\eqref{SINR}$, the achievable throughput per LEO is obtained as
\begin{align} \label{LEOrate}
R_{n} = \sum_{m=1}^M\sum_{u=1}^{U}\sum_{s=1}^{S} W_{s} \rho_{n,m,s} \cdot \log_{2}\left(1+\Gamma_{n,m,u,s}\right),
\end{align}
whilst the total system throughput is $R_{tot} = \sum_{n=1}^{N} R_n$. Our objective aims for maximizing the total throughput by considering policies of power $\boldsymbol{P}=\left\{ P_{n,m,u,s}|\forall n,m,u,s \right\}$, beam $\boldsymbol\phi =\left\{ \phi_{n,m,u}| \forall n,m,u \right\}$, and channel $\boldsymbol{\rho} =\left\{ \rho_{n,m,s}| \forall n,m,s \right\}$ with available resource constraints. Hence, the optimization problem can be formulated as
\begingroup
\allowdisplaybreaks
\begin{subequations} \label{P2}
\begin{align}
	\mathop{\max}_{\boldsymbol{P}, \boldsymbol\phi , \boldsymbol{\rho}} \quad & R_{tot} \label{obj}\\
\text{s.t. } 
	& 0 \leq \sum_{m=1}^M \sum_{u=1}^U \sum_{s=1}^S P_{n,m,u,s} \leq P_{leo} ,  &\forall {n}, \label{C1}\\
	& 0 \leq \sum_{u=1}^{U} \sum_{s=1}^{S} P_{n,m,u,s} \leq P_{beam} ,  &\forall {n,m}, \label{C2}\\
	& \sum_{n=1}^N \sum_{m=1}^M\phi_{n,m,u} \leq 1, &\forall {u} ,\label{C3}\\
	& \sum_{m=1}^M \rho_{n,m,s}\leq 1, &\forall {n,s} ,\label{C4}\\
	& 1 \leq \sum_{s=1}^S \phi_{n,m,u}\rho_{n,m,s} \leq S, &\forall {n,m,u} ,\label{C5}\\
	& \phi_{n,m,u}\in\{0,1\}, &\forall {n,m,u} ,\label{C6}\\
	& \rho_{n,m,s}\in\{0,1\}, &\forall {n,m,s} .\label{C7}
\end{align}
\end{subequations}
\endgroup
In $\eqref{C1}$ and $\eqref{C2}$, we constrain the per LEO transmit power and available power budget on each beam, respectively. In $\eqref{C3}$, each user can be served by at most one beam of an LEO. In $\eqref{C4}$, each sub-channel can only be assigned to a single illuminated beam. To guarantee service, in $\eqref{C5}$, we should allocate at least one sub-channel of a beam. $\eqref{C6}$ and $\eqref{C7}$ are generic binary set constraints. Due to the coupled continuous and discrete variables, it induces difficulty of a non-linear and non-convex problem $\eqref{P2}$, which is too complex to be resolved. Additionally, it becomes impractical for LEOs to estimate all user channels to acquire the exact throughput solutions in either centralized or distributed architecture, which will be addressed in the following section.

\begin{figure}
  \centering
    \includegraphics[width=3.2in]{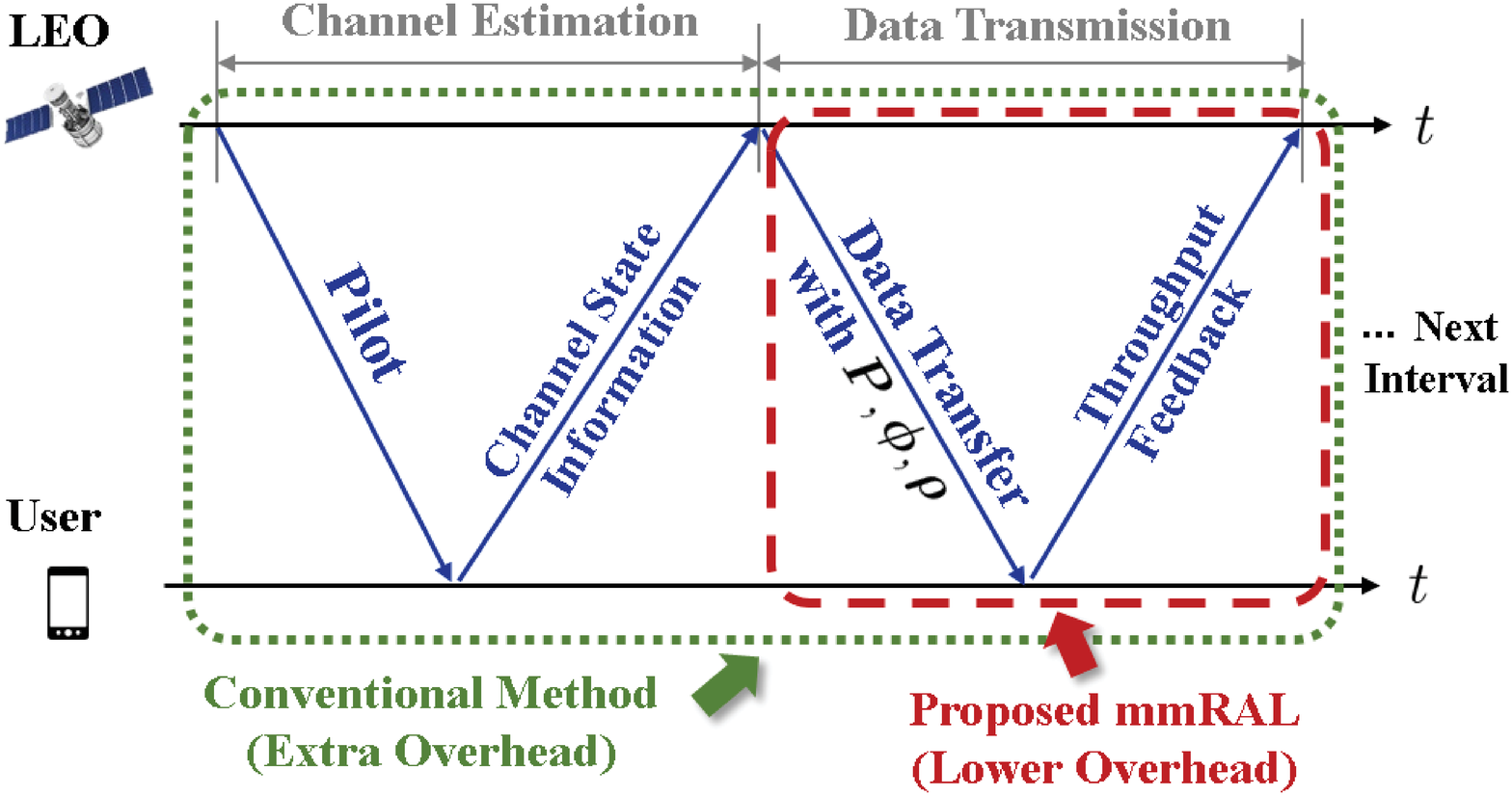}
  \caption{The framework containing estimation and transmission period. The proposed system requires only signalling of allocation and throughput feedbacks (red dotted box), whereas conventional protocol requires additional overhead of channel-pilot estimation and allocation (green dotted box).}
  \label{fig:method}
\end{figure}

\section{Proposed Hierarchical Multi-Agent Multi-Armed Bandit Resource Allocation for LEO Constellation (mmRAL)} \label{alg}

As shown in Fig. \ref{fig:method}, conventional transmission requires four-way handshake, including additional pilot and channel estimations. This potentially induces inaccurate estimation under high-mobility LEOs and considerably long-delay distance. Therefore, we propose a framework requiring no channel information but with only accumulated throughput feedbacks from users. Benefited by such mechanism, overhead is circumvented and the system can be adaptive to highly-dynamic conditions. To elaborate a little further, total rate $R_{tot}$ in $\eqref{obj}$ can be acquired in the ground gateway by collecting all dataset from LEOs, which is regarded as an implementable complexity. Without loss of generality, an LEO can acquire its individual performance metric of $R_n$ from its serving users. Therefore, LEOs should collaboratively assign the given resources of beams, power and channels to achieve high-throughput as well as low-interference.

The proposed mmRAL scheme is based on the designed hierarchical multi-agent MAB (MA-MAB) architecture, including \textit{macro}-agents (or so-called players) of LEOs and \textit{micro}-agents of resources. Note that different from the state-driven reinforcement learning, MA-MAB operates with only actions (or so-called arms) and rewards through exploitation and exploration. The LEO network consists of multi-agents of LEOs with global action set $\mathcal{A} =\left\lbrace \mathcal{A}_n=\{ \boldsymbol{P}_n, \boldsymbol{\phi}_n , \boldsymbol{\rho}_n  \} | \forall 1\leq n\leq N \right\rbrace$ and reward $\mathcal{R}=\left\lbrace \mathcal{R}_n = \{ R_{tot}, R_n \} | \forall 1\leq n\leq N \right\rbrace$. Therefore, each LEO $n$ will update its policy based on the designed update rule of MAB table $\mathcal{M}_{n}\left( \mathcal{A}_n^{t} \right)$ as
\begingroup
\allowdisplaybreaks
\begin{align}\label{mab_leo}
	&\mathcal{M}_{n}\left( \mathcal{A}_n^{t+1} \right) \leftarrow  \mathcal{M}_{n}\left( \mathcal{A}_n^{t} \right) \notag \\ & + \frac{1}{N\left(  \mathcal{A}_n^{t} \right)} \cdot \left[ R_{tot}^t + \gamma_n^t \max_{\mathcal{A}'_n} \Big( \mathcal{M}_{n}\left( \mathcal{A}'_n \right) - \mathcal{M}_{n}\left( \mathcal{A}_n^{t} \right) \Big)  \right],
\end{align}
\endgroup
where $t$ indicates the iteration index. The symbol $N\left(  \mathcal{A}_n^{t} \right)$ is the cumulated numbers of actions previously conducted, whilst $\gamma_n^t$ is the weight for scaling the potential optimal solution from $\max_{\mathcal{A}'_n} \mathcal{M}\left( \mathcal{A}'_n \right)$. Additionally, as can be observed in $\mathcal{A}_n$, quantization of actions is required due to continuous power set. This also induces an compellingly large searching space by jointly considering all three indicators. As a result, to reduce the computational complexity, we design a micro-scale MA-MAB by regarding resources as separate power/beam/channel-agents with their respective actions for the $n$-th LEO of $\mathcal{A}_{n,P}\!=\!\{ \boldsymbol{P}_n\}$, $\mathcal{A}_{n,\phi}\!=\!\{\boldsymbol{\phi}_n\}$, and $\mathcal{A}_{n,\rho}\!=\!\{\boldsymbol{\rho}_n\}$ and the shared reward of $\mathcal{R}_{n,P} \!=\! \mathcal{R}_{n,\phi} \!=\! \mathcal{R}_{n,\rho} \!=\!\{R_n\}, \forall 1\leq n \leq N$. Derived from $\eqref{mab_leo}$, the respective update policy of power/beam/channel-agents' MAB table can be represented as
\begingroup
\allowdisplaybreaks
\begin{align}\label{mab_micro}
	&\mathcal{M}_{n,\mathcal{X}} \left( \mathcal{A}_{n,\mathcal{X}}^{t+1} \right) \leftarrow  \mathcal{M}_{n,\mathcal{X}}\left( \mathcal{A}_{n,\mathcal{X}}^{t} \right)+ \frac{1}{N\left(  \mathcal{A}_{n,\mathcal{X}}^{t} \right)} \notag \\ & \cdot \left[ R_{n}^t + \gamma_{n,\mathcal{X}}^t \max_{\mathcal{A}'_{n,\mathcal{X}}} \Big( \mathcal{M}_{n,\mathcal{X}}\left( \mathcal{A}'_{n,\mathcal{X}} \right) - \mathcal{M}_{n,\mathcal{X}}\left( \mathcal{A}_{n,\mathcal{X}}^{t} \right) \Big)  \right],
\end{align}
\endgroup
where subscript $\mathcal{X}=\{P,\phi,\rho\}$ indicates the micro-agent of resources. We notice that the same iteration index is adopted in $\eqref{mab_leo}$ and $\eqref{mab_micro}$. This is because both sum and individual rates are acquired after users' feedbacks accordingly to the policy determined. To elaborate a little further, exploitation and exploration are the indispensable mechanism to prevent locally-optimal solution but keep searching the optimum, which can be designed based on greedy-based algorithm as
\begin{equation} \label{greedy}
\mathcal{A}_{n,\mathcal{X}} \!=\!\left\{  
\begin{aligned}  
& \text{Randomly select from }\mathcal{A}_{n,\mathcal{X}} , & \text{ if } rand()\leq \epsilon ,\\  
& {\arg\!\max}_{\mathcal{A}'_{n,\mathcal{X}}}  \mathcal{M}_{n,\mathcal{X}}\left( \mathcal{A}'_{n,\mathcal{X}} \right), &\quad \text{otherwise}, \\ 
\end{aligned}  
\right.
\end{equation}
where $rand()$ is the pseudo randomization generating an arbitrary value between $\left[0,1\right]$, and $\epsilon$ is the threshold for random action in greedy algorithm. The concrete algorithm is demonstrated in Algorithm \ref{codealg}. We consider that all LEOs are synchronized by a ground gateway. It is worth noting that all LEOs simultaneously perform micro-agent processing in Lines 7--9 in Algorithm \ref{codealg}. The rewards are then obtained after completion of policy determination and services from LEOs. Both micro- and macro-agents will update their MAB tables in Lines 12 and 13, respectively. Note that system and individual throughput are utilized for updating macro- and micro-agents' MAB tables, respectively. The proposed mmRAL scheme can be generically adapted to the condition of comparably stationery LEO constellations, i.e., coherent channels under low-mobility. To elaborate a little further, the proposed mmRAL requires no computational complexity and genuinely faster convergence benefited from considerably smaller searching spaces. The original problem has an exhaustive action space order $\mathcal{O} \left( |\boldsymbol{P}| \cdot 2^{ |\boldsymbol{\phi}|\cdot |\boldsymbol{\rho}| }  \right)$, whilst it requires a space order $\mathcal{O}\left( |\boldsymbol{P}| \cdot |\boldsymbol{\phi}| \cdot |\boldsymbol{\rho}| \right)$ for macro-agents updating $\eqref{mab_leo}$. However, proposed mmRAL has a moderate space order of $\mathcal{O}\left( \max( |\boldsymbol{P}|, |\boldsymbol{\phi}|, |\boldsymbol{\rho}| \right)$ from $\eqref{mab_micro}$ depending on the maximum space order of resources.

\begin{figure}
  \centering
  \includegraphics[width=3.5in]{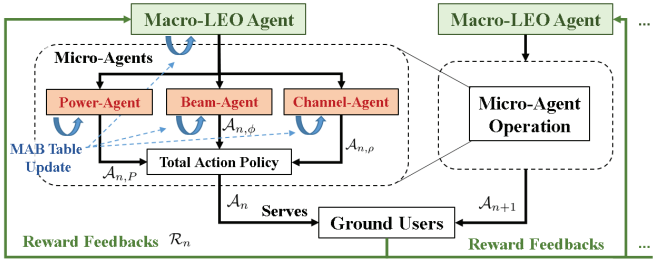}
  \caption{The proposed mmRAL scheme. Hierarchical agents include collaborative multiple macro-agents of LEOs and competitive micro-resource-agents, i.e., power/beam/channel-agents. Note that only throughput is required for learning appropriate allocation actions.}
  \label{fig:Algorithm}
\end{figure}

%-------------------------------
 \begin{algorithm}[!tb]
    \footnotesize
        \caption{\small Proposed mmRAL Scheme for LEO Constellation}
        \SetAlgoLined
        \DontPrintSemicolon
        \label{codealg}
        \begin{algorithmic}[1]
        \STATE {\bf Initialization:} Iteration index $t=0$, thresholds of $\epsilon$, weight $\gamma_{n,\mathcal{X}}^t, \forall t$, available power of $P_{leo}$, $P_{beam}$
        \STATE Set empty table $\mathcal{M}_{n,\mathcal{X}} \left( \mathcal{A}_{n,\mathcal{X}}^{t} \right)= \emptyset, \forall \mathcal{X}=\{P,\phi,\rho\}, 1\leq n\leq N$
        \STATE Ground gateway synchronizes all LEOs
        \WHILE{LEO constellation is operating}
        	\STATE LEO moves while channel varying with $t\leftarrow t+1$
        		\STATE Simultaneous \textbf{Exploration--Exploitation} for all LEOs $n=1,...,N$
        		\STATE Conduct $\eqref{greedy}$ for power-agent $n$'s power allocation action as $\mathcal{A}_{n,P}^t$ constrained by $\eqref{C1}$, $\eqref{C2}$
        		\STATE Perform $\eqref{greedy}$ for beam-agent $n$'s illuminated beam action as $\mathcal{A}_{n,\phi}^t$ fulfilling $\eqref{C3}$, $\eqref{C5}$, $\eqref{C6}$
        		\STATE Execute $\eqref{greedy}$ for channel-agent $n$'s sub-channel assignment action as $\mathcal{A}_{n,\rho}^t$ satisfying $\eqref{C4}$, $\eqref{C5}$, $\eqref{C7}$
             \STATE All LEOs serve ground users using the determined policy of $\mathcal{A}_{n,\mathcal{X}}^t, \forall \mathcal{X}=\{P,\phi,\rho\}$
            \STATE Obtain the individual LEO rate $R_{n}$ and total rate $R_{tot}$
            \STATE Micro-agents update their MAB tables $\mathcal{M}_{n,\mathcal{X}} \left( \mathcal{A}_{n,\mathcal{X}}^{t} \right)$, $\forall \mathcal{X}=\{P,\phi,\rho\}$ using $\eqref{mab_micro}$ with the determined actions
            \STATE LEO macro-agents update their MAB tables $\mathcal{M}_{n}\left( \mathcal{A}_n^{t} \right)$ using $\eqref{mab_leo}$ \\ with assembled policy $\mathcal{A}_n^t= \bigcup_{\mathcal{X}\in \{P,\phi,\rho\}} \mathcal{A}_{n,\mathcal{X}}^{t} $
        \ENDWHILE
      \end{algorithmic}
    \end{algorithm}

\begin{table}[!t]
\footnotesize
\centering
    \caption {Parameters Setting}
        \begin{tabular}{ll}
            \hline
            System Parameter & Value \\ \hline \hline  
            Beam cells per LEO $M$ & $19$ \\ 
            Number of sub-channels $S$ & $30$ \\
            Illuminated beam cells & $15$ \\
            LEO serving radius & $500$ km \\
            Inter-LEO satellite distance & $\{250,500\}$ km \\
            Operating band $f_c$ & $28$ GHz (Ka-band) \\
            System bandwidth $W$ & 240 MHz \\
            LEO/Beam transmit power ${P}_{leo}$, ${P}_{beam}$ & $60$, $40$ dBm \\
            Noise power spectral density ${N}_0$ & $-174$ dBm/Hz \\
            Doppler shift compensation $\eta$ & $-120$ dBW \\
            Antenna aperture & $10c/f_c$ rad \\
            Transmitter antenna aperture gain $G_t$ & $50$ dBi \\
            Receiver gain ${G}^R_u$ & $15$ dBi \\
            LEO channel iterations $t$ & $\left[0, 20000\right]$ \\
            Greedy threshold $\epsilon$ & $0.2$ \\
            Weight for MAB table $\gamma_{n}^t$, $\gamma_{n,\mathcal{X}}^t$ & $0.15, 0.15$ \\
            LEO constellation & Walker \\
            Velocity of LEOs/users & $8$ km/s, $20$ m/s \\
            \hline
        \end{tabular} \label{table1}
    \end{table}    

\section{Performance Evaluation} \label{sim}

	In this section, we present the performance of proposed mmRAL scheme in multi-LEOs serving multiple ground users. The pertinent parameters are listed in Table \ref{table1} referenced from \cite{RABPC, LEOmovement, benchmark2}. Note that we define beam radius as $BR$, which is calculated based on the geometric relationship \cite{LEOpromising2} between predefined $M=19$ beam cells and LEO positions with height around $h_n=1000$ km. Users are randomly and uniformly distributed in all LEO serving coverage. We also constrain that the illuminated beam cells should be fewer than all beam cells for resiliently and efficiently using ample resources. Other used LEO channel-related parameters can be found in \cite{channel1,channel2}.

\begin{figure}[t]
  \centering
  \includegraphics[width=3in]{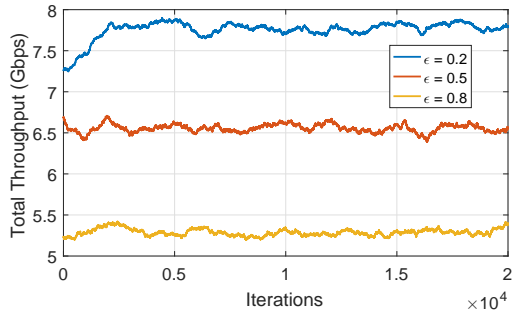}
  \caption{Convergence of proposed mmRAL scheme with respect to different $\epsilon$-greedy parameters $\epsilon=\{0.2, 0.5, 0.8\}$.}
  \label{fig1}
\end{figure}

In Fig. \ref{fig1}, we evaluate the convergence of mmRAL with $\epsilon = \{0.2, 0.5, 0.8\}$. With larger values of $\epsilon$, we have a higher tendency of selecting randomized actions leading to random resource allocation, which is revealed with $\epsilon=0.8$ having around $5.3$ Gbps throughput. When $\epsilon = 0.5$, half duration of the LEO operation will conduct suitable learned actions from random exploration. However, both curves with $\epsilon=\{0.5,0.8\}$ demonstrate the non-convergence due to highly-random selections. On the contrary, with an appropriate setting of $\epsilon=0.2$, it can strike a tradeoff between exploration and exploitation asymptotically converging to the highest throughput performance of around $7.9$ Gbps, which is the candidate setting in the remaining evaluations.

\begin{figure}[!ht]
		\centering
		\subfigure[]{
		\includegraphics[width=1.7in]{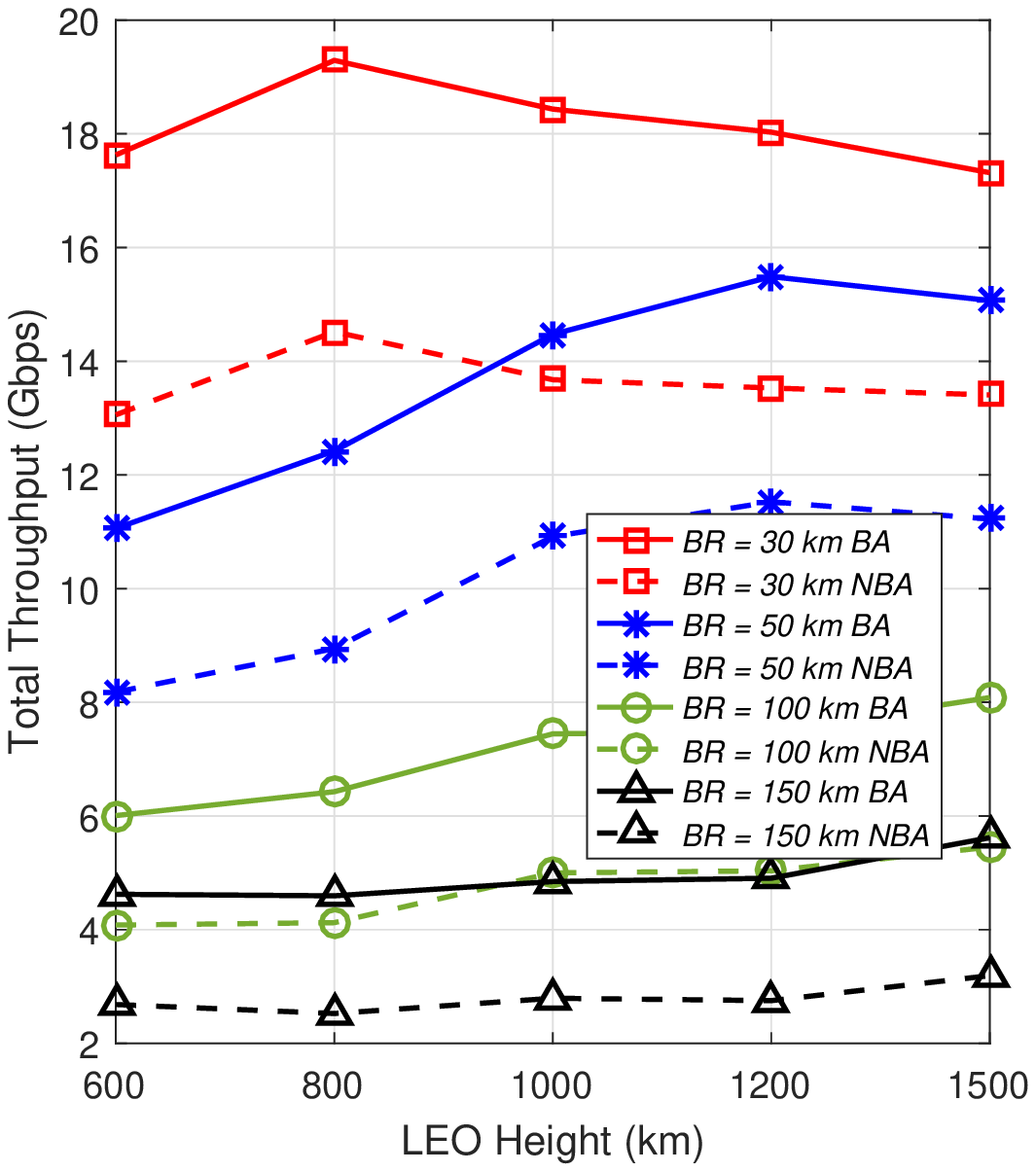} \label{fig2-1}}
		\subfigure[]{
		\includegraphics[width=1.7in]{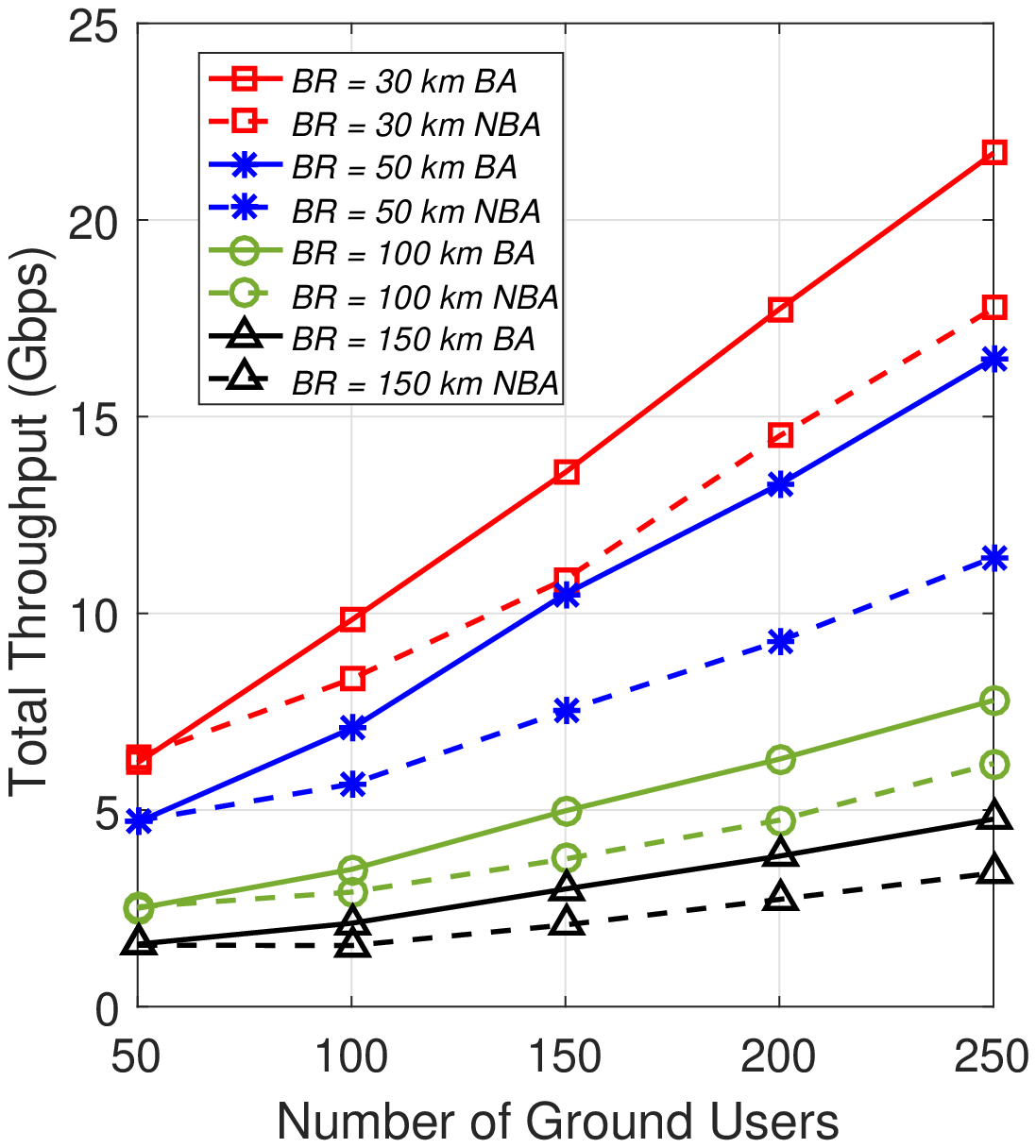} \label{fig2-2}}
		\caption{Performance of proposed mmRAL with beam allocation, BA and non-BA, in terms of different (a) LEO heights and (b) numbers of serving ground users.}
		\label{fig2}
\end{figure}	

In Fig. \ref{fig2}, we elaborate the effects of different beam cell radius $BR=\{30,50,100,150\}$ km with beam allocation (BA) and non-beam allocation (NBA), i.e., full-beam illumination. In Fig. \ref{fig2-1}, we consider different LEO heights of $h_n=\{600, 800, 1000, 1200, 1500\}$ km. It can be seen that BA has around $1.3$ to $1.5$ times improved throughput than that of NBA. With increment of deployment height, it has higher throughput due to lower interference from high pathloss. Moreover, throughput increases with smaller $BR$ having comparatively higher beam-gains from narrower beamwidths. However, optimal deployment heights take place when $h_n=800$ km for $BR=30$ km and $h_n=1200$ km for $BR=50$ km. This is because pathloss is predominant with more resources required to generate multi-beams for high coverage probability. As shown in Table \ref{table2}, it validates that massive antennas are required to generate multi-narrow-beams with smaller $BR$ and higher deployment $h_n$, which leads to insufficient assigned resources and declined throughput as explained in Fig. \ref{fig2-1}.

 As depicted in Fig. \ref{fig2-2}, we consider different numbers of ground users of $U=\{50,100,150,200,250\}$. Higher throughput is obtainable because of more resilient resource allocation policy with the increment of serving users. By comparing BA and NBA schemes, larger throughput difference is revealed when $U=250$ thanks to interference alleviation with high degree of freedom for selecting candidate users with better channel qualities with limited resources. Furthermore, Table \ref{table3} demonstrates the outage probability indicating that user throughput falls below the predefined threshold of $50$ Mbps. Due to constrained illuminated beam cells, more users will induce higher outage but also offer opportunities of mitigating interference for serving users at the timeslot. However, NBA with full-beam illumination cannot generate low-interfered beams, which results in higher outage probability than BA.

% homogeneous heteregeneous
\begin{figure}[t]
  \centering
  \includegraphics[width=2.8in]{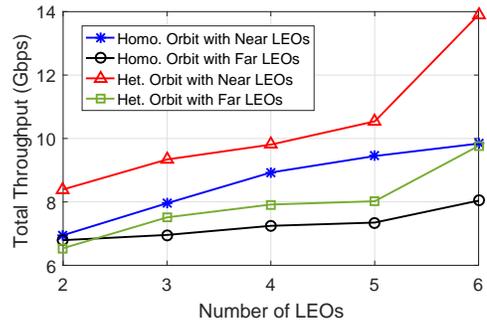}
  \caption{Performance with far/near LEOs with respective longer/shorter inter-LEO satellite distances under homogeneous and heterogeneous LEO orbit topologies.}
  \label{fig3}
\end{figure}

\begin{figure}[t]
  \centering
  \includegraphics[width=2.8in]{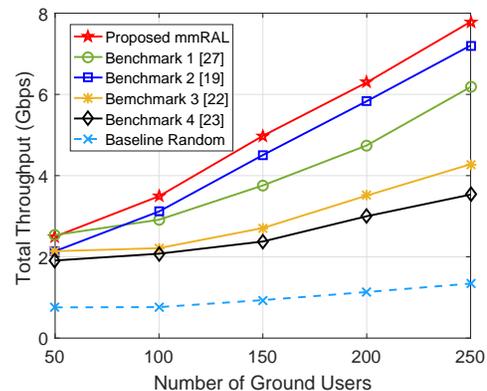}
  \caption{Performance comparison of proposed mmRAL scheme with existing benchmarks in open literature and baseline of random allocation.}
  \label{fig4}
\end{figure}

\begin{table*}[t] 
\scriptsize
\centering
\caption{Required Number of Antennas on LEO}
\renewcommand{\arraystretch}{1.1}
\label{table2}
\begin{tabular}{|c||c|c|c|c|c|c|c|c|c|c|}	
	\cline{1-11}
	\multirow{2}{*}{$BR$} & \multicolumn{2}{c|}{$h$=600 km} & \multicolumn{2}{c|}{$h$=800 km} & \multicolumn{2}{c|}{$h$=1000 km} & \multicolumn{2}{c|}{$h$=1200 km} & \multicolumn{2}{c|}{$h$=1500 km}\\ 
	\cline{2-11}
	& BA & NBA & BA & NBA & BA & NBA & BA & NBA & BA & NBA \\ \hline 
	30 km & 15 & 19 & 60 & 76 & 60 & 76 & 60 & 76 & 135 & 171 \\ \hline
	50 km  & 15 & 19 & 15 & 19 & 15 & 19 & 60 & 76 & 60 & 76 \\ \hline
	100 km & 15 & 19 & 15 & 19 & 15 & 19 & 15 & 19 & 15 & 19 \\ \hline
 	150 km & 15 & 19 & 15 & 19 & 15 & 19 & 15 & 19 & 15 & 19 \\ \hline
\end{tabular}
\end{table*}

\begin{table*}[t] 
\scriptsize
\centering
\caption{Outage Probability}
\renewcommand{\arraystretch}{1.1}
\label{table3}
\begin{tabular}{|c||c|c|c|c|c|c|c|c|c|c|}  
	\cline{1-11}
	 \multirow{2}{*}{$BR$} & \multicolumn{2}{c|}{$U=50$ users} & \multicolumn{2}{c|}{$U=100$} & \multicolumn{2}{c|}{$U=150$} & \multicolumn{2}{c|}{$U=200$} & \multicolumn{2}{c|}{$U=250$}\\ 
	\cline{2-11}
	& BA & NBA & BA & NBA & BA & NBA & BA & NBA & BA & NBA \\ \hline 
	30 km & 4\% & 4\% & 11\% & 17\% & 14.67\% & 20.67\% & 18\% & 23\% & 18.8\% & 25.6\% \\ \hline
	50 km  & 16\% & 18\% & 37\% & 58\% & 40\% & 60\% & 40\% & 69.5\% & 40.4\% & 65.2\% \\ \hline
	100 km & 70\% & 70\% & 80\% & 91\% & 80.67\% & 94\% & 80\% & 93\% & 79.6\% & 90\% \\ \hline
 	150 km & 82\% & 84\% & 90\% & 96\% & 92\% & 96.67\% & 90\% & 96.5\% & 89.6\% & 96\% \\ \hline
\end{tabular}
\end{table*}

In Fig. \ref{fig3}, we evaluate the homogeneity (Homo.) and heterogeneity (Het.) of LEO orbital topologies. Homo. orbit means that LEOs are moving within the same orbit plane, whilst Het. orbit indicates different orbits for LEOs. Note that we consider inter-LEO satellite distance of $\{250,500\}$ km for far and near LEOs, respectively. At most $2$ LEOs are deployed in the same orbit for Het. orbit case. We can know from Fig. \ref{fig3} that more LEOs offer higher throughput due to more obtainable resources. Het. orbit achieves higher throughput than that of Homo. one because of less overlapped serving regions inducing compellingly lower interference. Interestingly, the performance for topology of near LEOs outperforms the case of far LEOs. This phenomenon takes place due to the fact that neighboring LEOs attempt to illuminate fewer elite beams with more assigned resources under better channel quality.

In Fig. \ref{fig4}, numerical results are provided by comparing different benchmarks and baseline (random allocation) with different numbers of users. \textbf{Benchmark 1} \cite{benchmark1} considers only channel allocation and power control with full-beam illumination, whilst \textbf{Benchmark 2} \cite{benchmark2} only cares for beam/sub-channel assignment with full power utilization. The work of \textbf{Benchmark 3} in \cite{benchmark3} considers pure LEO-channel allocation, whereas \textbf{Benchmark 4} in \cite{benchmark4} focuses on optimizing power. All the above-mentioned works adopt reinforcement learning-based algorithm, which requires additional channel information. Benchmarks 1--4 perform joint decision with comparatively larger action searching spaces, which potentially converges more slowly. Moreover, they did not consider severe mobility-aware interference. Accordingly, the proposed mmRAL scheme is capable of intelligently learning an appropriate allocation policy by jointly considering power/beam/channel resources, which alleviates inter-beam, inter-LEO and Doppler-induced interferences. mmRAL can achieve the highest throughput and higher rate difference with more serving ground users compared to the baseline as well as the existing benchmarks in open literature.

\section{Conclusion} \label{con}
In this paper, we have conceived a multi-LEO constellation with the two-way transmission framework under no channel knowledge. Comprehensive interference of inter-beam, inter-LEO and mobility-aware interferences are taken into account. The mmRAL scheme is designed with the concept of hierarchical architecture and MA-MAB. Collaborative macro-LEO agents mandate the respective micro-agents of power/beam/channel resources to strike a compelling tradeoff between exploitation and exploration in a greedy manner. Simulations have demonstrated the potential benefits of the proposed mmRAL in terms of optimal deployment of LEOs in different beam radius and Homo./Het. orbital topologies, as well as supporting various numbers of LEOs and users in a wider coverage. Thanks to the intelligent learning for resilient resource allocation, the proposed mmRAL scheme is capable of achieving the highest throughput performance with a moderate complexity order by comparing to the existing benchmarks in open literature.

\setstretch{0.85}
\bibliographystyle{IEEEtran}
\bibliography{IEEEabrv}

\end{document}